\documentclass{article}

\usepackage{microtype}
\usepackage{graphicx}
\usepackage{subfigure}
\usepackage{booktabs} 

\usepackage{hyperref}

\usepackage[accepted]{icml2024}

\usepackage{amsmath}
\usepackage{amssymb}
\usepackage{mathtools}
\usepackage{amsthm}

\usepackage[capitalize,noabbrev]{cleveref}

\theoremstyle{plain}
\newtheorem{theorem}{Theorem}[section]

\theoremstyle{definition}
\newtheorem{definition}[theorem]{Definition}

\theoremstyle{remark}

\usepackage[textsize=tiny]{todonotes}

\usepackage{multirow}
\usepackage{enumitem}
\usepackage{makecell}

\icmltitlerunning{Higher Layers Need More LoRA Experts}

\begin{document}

\twocolumn[
\icmltitle{Higher Layers Need More LoRA Experts}


\icmlsetsymbol{equal}{*}

\begin{icmlauthorlist}
\icmlauthor{Chongyang Gao}{nu}
\icmlauthor{Kezhen Chen}{comp}
\icmlauthor{Jinmeng Rao}{comp}
\icmlauthor{Baochen Sun}{comp}
\icmlauthor{Ruibo Liu}{DM}
\icmlauthor{Daiyi Peng}{DM}
\icmlauthor{Yawen Zhang}{comp}
\icmlauthor{Xiaoyuan Guo}{comp}
\icmlauthor{Jie Yang}{comp}
\icmlauthor{VS Subrahmanian}{nu}

\end{icmlauthorlist}

\icmlaffiliation{nu}{Department of Computer Science, Northwestern University, Evanston, IL}
\icmlaffiliation{comp}{Mineral Research, USA}
\icmlaffiliation{DM}{Google DeepMind, USA}

\icmlcorrespondingauthor{Chongyang Gao}{chongyanggao2026@u.northwestern.edu}
\icmlcorrespondingauthor{Kezhen Chen}{kzchen0204@gmail.com}
\icmlcorrespondingauthor{VS Subrahmanian}{vss@northwestern.edu}

\icmlkeywords{Mixture of Expert, Parameter-efficient Learning, LoRA, LLM, Instruction Tuning}

\vskip 0.3in
]



\printAffiliationsAndNotice{}  

\begin{abstract}
Parameter-efficient tuning (PEFT) techniques like low-rank adaptation (LoRA) offer training efficiency on Large Language Models, but their impact on model performance remains limited. Recent efforts integrate LoRA and Mixture-of-Experts (MoE) to improve the performance of PEFT methods. Despite promising results, research on improving the efficiency of LoRA with MoE is still in its early stages. Recent studies have shown that experts in the MoE architecture have different strengths and also exhibit some redundancy. Does this statement also apply to parameter-efficient MoE? In this paper, we introduce a novel parameter-efficient MoE method, \textit{\textbf{M}oE-L\textbf{o}RA with \textbf{L}ayer-wise Expert \textbf{A}llocation (MoLA)} for Transformer-based models, where each model layer has the flexibility to employ a varying number of LoRA experts. We investigate several architectures with varying layer-wise expert configurations. Experiments on six well-known NLP and commonsense QA benchmarks demonstrate that MoLA achieves equal or superior performance compared to all baselines. We find that allocating more LoRA experts to higher layers further enhances the effectiveness of models with a certain number of experts in total. With much fewer parameters, this allocation strategy outperforms the setting with the same number of experts in every layer. This work can be widely used as a plug-and-play parameter-efficient tuning approach for various applications. The code is available at \url{https://github.com/GCYZSL/MoLA}.
\end{abstract}

\section{Introduction}
\label{sec:intro}

Large Language Models (LLMs) have shown impressive proficiency and transfer learning capabilities across a variety of tasks and domains \citep{palm, zhang2023llama, gemini, mixtral, med-palm,judge-lm,fpf}. However, modern LLMs fine-tuning demands huge computational resources due to the vast number of parameters. To mitigate this issue, the research community is increasingly focusing on parameter-efficient fine-tuning (PEFT) methods to dramatically reduce training costs, such as p-tuning \citep{ptuning} or low-rank adaption (LoRA) \citep{lora}. Despite its training efficiency, PEFT methods' performance in fine-tuning LLMs is still limited.

Recent studies show that combining PEFT with the Mixture-of-Experts (MoE) becomes a promising recipe for leveraging MoE in a parameter-efficient fashion \citep{cohereloramoe,moelora,fudanloramoe}, providing impressive performance. Most of these methods apply MoE on LoRA, called LoRA-MoE. For Transformer models~\citep{transformer}, LoRA learns a pair of low-rank matrices as an adapter for a given dense linear layer, effectively modifying the layer's behavior without substantial change to the original model parameters. Instead of learning one pair of low-rank matrices, LoRA-MoE learns multiple pairs of low-rank matrices, called \textit{LoRA experts}, and a router to compute the weights of each expert for inputs. During the LLM fine-tuning phase, pre-trained weights of dense layers remain fixed, while LoRA experts and the router are trained to adapt the pre-trained weights. While the initial results are promising, the research into achieving more efficient and effective integration is still in its infancy.

Moreover, recent studies in the MoE analysis indicate that the use of many experts may be redundant due to representational collapse or learned routing policy overfitting \citep{moeanalysis1, moeanalysis2}. More experts in a layer may cause the representation to overfit the training data, as the data is processed in a more fine-grained manner. This insight leads us to think about how many experts could be more suitable for different layers in the Transformer model, motivating us to explore two questions. 

\textit{(i) Are there any redundant experts in parameter-efficient MoE? (ii) What strategy should be used to allocate the number of LoRA experts in each layer?}

To address these questions, we introduce a \emph{new} parameter-efficient MoE approach, \textit{\textbf{M}oE-L\textbf{o}RA with \textbf{L}ayer-wise Expert \textbf{A}llocation (MoLA)}, combining LoRA and MoE with layer-wise expert allocation. Users can flexibly assign a different number of LoRA experts to each Transformer layer. We study several typical architectures with different layer-wise expert configurations. Using a fixed number of experts in total, we allocate them differently, with either lower layers or higher layers having more experts. We conduct experiments on six benchmarks including NLP and commonsense question-answering tasks to demonstrate the effectiveness of our MoLA approach under different configurations.

\emph{Key Findings:} Our extensive experiments reveal that experts in lower layers are more similar to each other and thus exhibit more redundancy. With a fixed number of experts, more LoRA experts should be allocated to the higher layers of the Transformer model to enhance its effectiveness. Our key contributions are: 

\begin{itemize}
\item We present a new parameter-efficient MoE method, MoLA, with flexible layer-wise expert allocation, on the Transformer model. MoLA integrates LoRA and MoE and introduces flexibility to assign different numbers of experts to different Transformer layers, reducing expert redundancy and diversifying information granularity. MoLA is a plug-and-play approach and can be applied to diverse tasks.

\item We study several MoLA variants on an LLM, each with different layer-wise expert configurations. Experiments on six benchmarks show that all MoLA configurations significantly outperform other PEFT baselines, showing the efficacy of our approach.

\item We further compare each layer-wise configuration of expert allocation.
\emph{Overall, the configuration, that has more LoRA experts in the higher layers and fewer in the lower layer, outperforms all other configurations.} Such specialized expert allocation configuration enables models to achieve enhanced performance vis-a-vis other configurations, even with much fewer parameters, demonstrating improved scalability.

\item Our comprehensive analysis shows that experts in lower layers are more similar than those in higher layers and thus have higher redundancy, providing insights into our observations.
\end{itemize}

\begin{figure*}[htb]
\begin{center}
\centerline{\includegraphics[width=1.5\columnwidth]{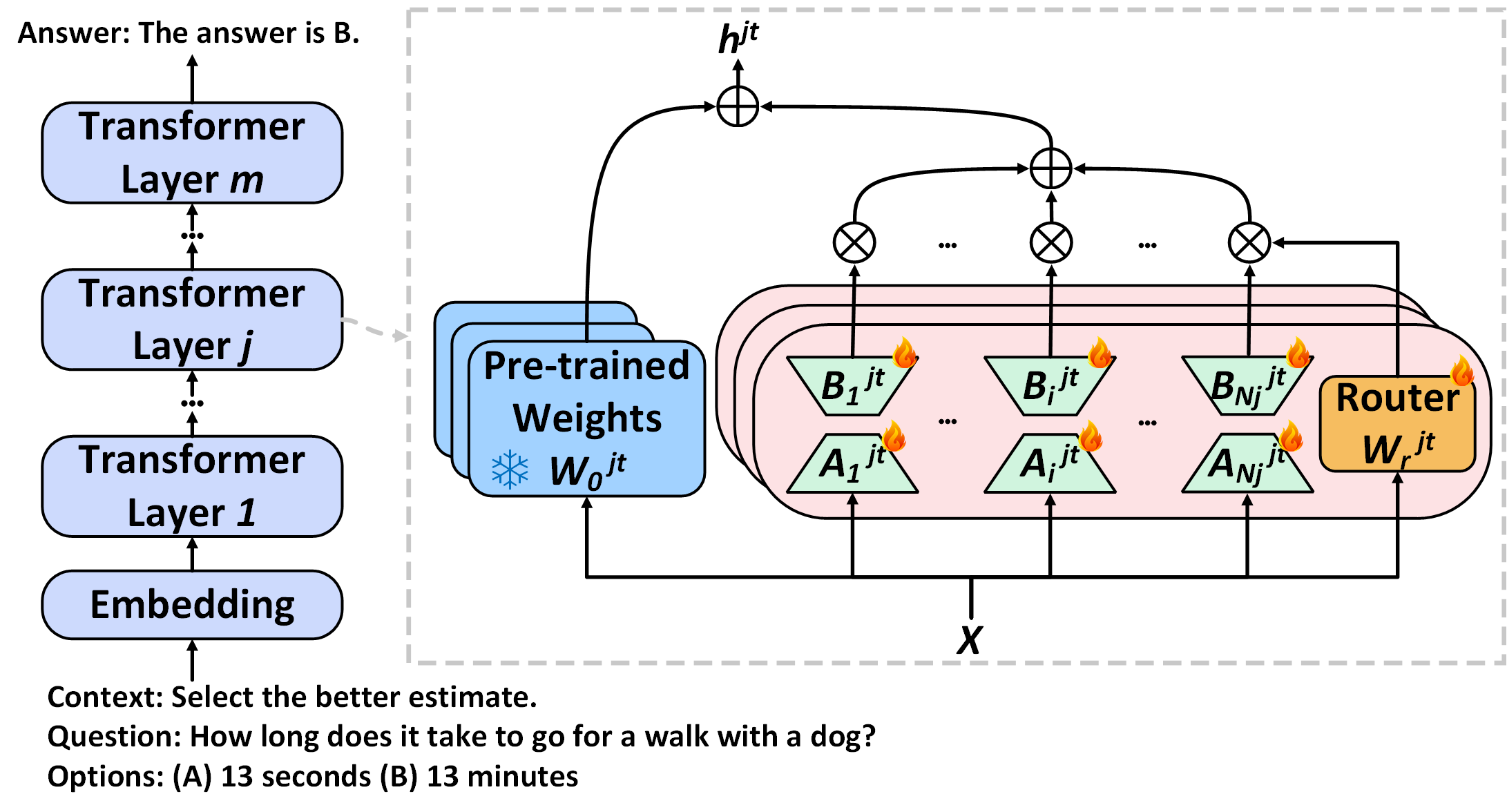}}
\caption{The overview of MoLA architecture. MoLA applies LoRA-MoE on a pre-trained Transformer model with layer-wise expert allocation. Each layer employs a different number of experts. During training, the pre-trained weights are freeze and only LoRA experts are tuned as the adapters on the weights.}
\label{fig1:MoLA}
\end{center}
\end{figure*}

\section{Preliminaries}
\label{sec:preliminaries}

We first briefly review MoE and LoRA before describing our MoLA framework.

\paragraph{Mixture of Experts} The MoE architecture \citep{moe} applies sparse sub-modules, called experts, to various inputs via a router module. The router module intelligently employs different experts for different types of inputs, thus scaling up model parameters with a constant computational cost. MoE has shown promising effectiveness on the Transformer model \citep{moe}. The MoE layer consists of $N$ identical and independent feed-forward neural networks $\{E\}_{i=1}^{N}$ as experts. The router is a gating function with a trainable weight matrix $W_{r}$. Given an input $x$, the router maps $x$ to an $N$-dimensional vector, which corresponds to the number of experts. The router uses a softmax function to compute a probability distribution of the weights of outputs from the expert networks. Following standard MoE architectures, only the top $K$ experts, determined by the router, are chosen for the computation. Additionally, an auxiliary loss, called load balancing loss, is used on each MoE layer to promote a balanced top-k selection by pushing the router to have equitable workload distribution among experts. Equation~\ref{eq1:moe} mathematically represents the MoE layer where $y$ is the output embedding from the MoE layer. With fine-tuning, different experts focus on processing different types of information or tasks and thus provide finer granularity. 

\begin{equation}
y = \sum_{i=1}^{K} \frac{\textrm{TopK}(\textrm{Softmax}(W_{r} x), K)_i}{\sum_{i=1}^{K}\textrm{TopK}(\textrm{Softmax}(W_{r} x), K)_i} E_i(x)
\label{eq1:moe}
\end{equation}

\paragraph{LoRA} LoRA is a popular parameter-efficient tuning approach that is widely used in LLM fine-tuning\cite{lora, adalora, qlora}. LoRA leverages low-rank matrix decomposition of pre-trained weight matrices to significantly reduce the number of training parameters. Given a pre-trained linear layer with a weight matrix $W_0 \in \mathbb{R}^{d_{q}\times d_{p}}$, LoRA creates two low-rank trainable matrices $A$ and $B$, where $A \in \mathbb{R}^{d_{q}\times r}$, $B \in \mathbb{R}^{r\times d_{p}}$, and $r \ll min(d_{q}, d_{p})$. Thus, the dimension of $ABx$ equals the dimension of $W_{0}x$. Equation~\ref{eq2:lora} mathematically describes this process and the output of LoRA is $h$. During training, $W_0$ is frozen and does not receive gradient updates, while $A$ and $B$ are updated.
\begin{equation}
h = W_{0}x + \triangle Wx = W_{0}x + ABx
\label{eq2:lora}
\end{equation}

The matrix $A$ is initialized with a random Gaussian distribution and matrix $B$ is initialized to zero. The initialization results in the same outputs as the original pre-trained model. When fine-tuning LLMs, the LoRA approach can be applied to all the linear layers in the Transformer model or its variants. Compared with tuning the original weight matrix, LoRA dramatically reduces the number of training parameters while keeping reasonable performance.


\section{MoE-LoRA with Layer-wise Allocation}
\label{sec:MoLA}
Combining MoE and LoRA has shown promising results \citep{cohereloramoe,moelora,fudanloramoe}. However, most such efforts only replace experts with LoRA adapters under the MoE framework, and each layer has a fixed number of experts. Thus, some shortcomings of MoE may persist in these methods. For instance, experts in MoE may be redundant due to representational collapse or learned routing policy overfitting \citep{moeanalysis1, moeanalysis2}. Inspired by this insight, we argue that the number of LoRA experts need \emph{not} be the same across all Transformer layers.

We thus introduce a novel parameter-efficient tuning approach, called MoE-LoRA with Layer-wise Allocation (MoLA), which combines LoRA and MoE techniques with smart layer-wise expert allocation. As most LLMs use Transformer-based architectures, we study how MoLA should be applied to the Transformer model. Instead of allocating the same number of experts to all layers of the Transformer, MoLA uses different numbers of experts on different layers. In this section, we first describe the details of our 
architecture and then propose several layer-wise expert allocations based on different assumptions.

\subsection{The MoLA Architecture}
\label{sec:MoLA_architecture}
MoLA integrates LoRA adapters into the MoE framework so each layer may have a different number of experts. When training a pre-trained LLM with LoRA, instead of decomposing each weight matrix of a dense linear layer into a pair of low-rank matrices, we create \emph{multiple} pairs of low-rank matrices --- each pair is called a LoRA expert. A router module is learned to route each input token to different LoRA experts.
Given a Transformer model with $m$ layers, we allocate $N_j$ experts for layer $j$ and have $\sum_{j=1}^{m} N_{j}$ experts in total.
Specifically, given a pre-trained weight matrix $W_0^{jt} \in \mathbb{R}^{d_{q}\times d_{p}}$ from the module $t$ in layer $j$, we create $N_j$ pairs of low-rank matrices $\{A^{jt}\}_{i=0}^{N_j}, \{B^{jt}\}_{i=0}^{N_j}$. As in the case of LoRA, each matrix $A_{i}^{jt}$ is initialized from a random Gaussian distribution. We set $B_{i}^{jt}$ to zero, where $A_{i}^{jt} \in \mathbb{R}^{d_{q}\times r}$, $B_{i}^{jt} \in \mathbb{R}^{r\times d_{p}}$, and $r \ll min(d_{q}, d_{p})$. Then, a router $S_{i}^{jt}$ with a trainable weight matrix $W_{r}^{jt} \in \mathbb{R}^{d_{q}\times N_j}$ is used to specify different LoRA experts for the input $x$. As in the original MoE, MoLA selects the top $K$ experts for computation and applies the load balancing loss on each layer. Figure~\ref{fig1:MoLA} shows an overview of the architecture. The mathematical representation is: 
\begin{equation}
S_{i}^{jt}(x) = \frac{\textrm{TopK}(\textrm{Softmax}(W_{r}^{jt} x), K)_i}{\sum_{i=1}^{K}\textrm{TopK}(\textrm{Softmax}(W_{r}^{jt} x), K)_i}
\label{eq3:router}
\end{equation}

\begin{equation}
h^{jt} = W_{0}^{jt}x + \sum_{i=1}^{K} S_{i}^{jt}(x) A_{i}^{jt}B_{i}^{jt}x
\label{eq4:MoLA}
\end{equation}

Eq.~\ref{eq3:router} represents the router with the input $x$ and Eq.~\ref{eq4:MoLA} mathematically shows the LoRA expert in MoLA, where $h^{jt}$ is the output embedding. This MoLA architecture provides the flexibility to modify the number of experts for each Transformer layer. The next section addresses the question of how experts should be allocated in each layer.

\subsection{Configurations of Layer-wise Expert Allocation}
\label{sec:MoLA_configurations}
\begin{figure}[tb]
\begin{center}
\centerline{\includegraphics[width=\columnwidth]{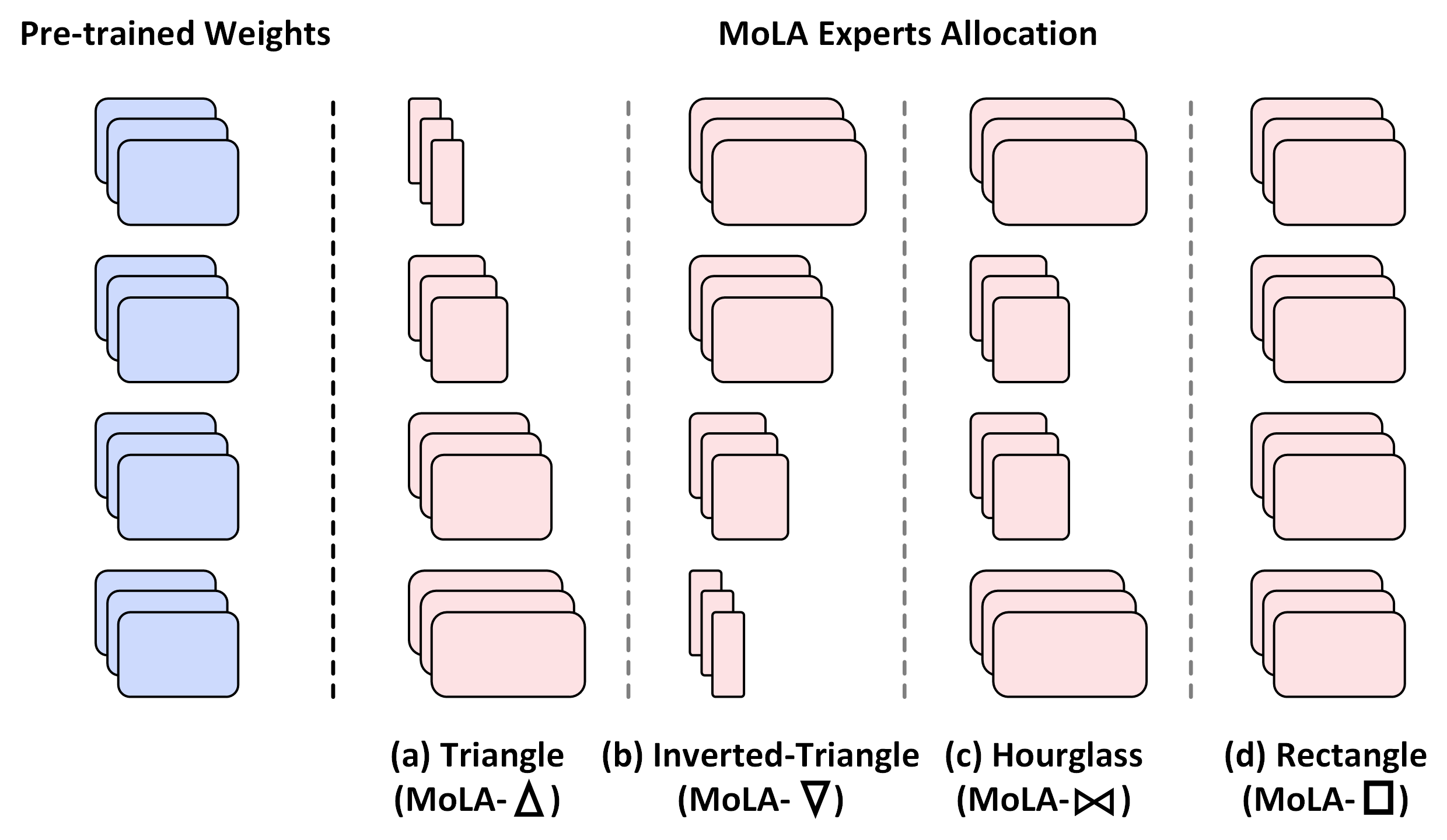}}
\caption{Four types of layer-wise expert allocations of MoLA.}
\label{fig2:MoLA-types}
\end{center}
\vskip -0.3in
\end{figure}

MoE is similar to an ensemble method with multiple experts learning fine-grained information. Layers with more experts have stronger fitting capabilities. One intuition is that we should allocate more experts to layers that are required to process diverse edge cases and fine-grained information. To study how LoRA experts should be allocated in each Transformer layer, we propose four types of layer-wise expert configurations based on different assumptions. Figure~\ref{fig2:MoLA-types} visualizes the overview of these four configurations.  Section~\ref{sec:experiments} describes detailed experiments to compare these configurations.

\paragraph{MoLA Triangle (MoLA-$\triangle$)}
Many studies have analyzed layer-wise representations of Transformer models. Generally, lower layers learn more token-level features, such as word meaning, syntax, or grammar, while higher layers capture more abstract, high-level information. As token-level information is subtle and diverse, one assumption is that token-level information may require more experts to distinguish fine-grained meaning while high-level information may require fewer experts for generalization. Our MoLA Triangle (MoLA-$\triangle$) architecture is based on this assumption and allocates experts in a ``triangle" shape: lower layers have more experts than higher layers. 

\paragraph{MoLA Inverted-Triangle (MoLA-$\triangledown$)} 
Unlike MoLA-$\triangle$, another assumption is that more experts for processing token-level information may introduce redundancy during the information processing. As higher layers learn more abstract and high-level information, and these features are used for downstream tasks, they may require more experts. More experts may enhance the architecture to process complicated problems by leveraging experts to learn fine-grained and task-specific patterns. Based on this intuition, we design the MoLA Inverted-Triangle (MoLA-$\triangledown$) configuration where lower layers are allocated fewer experts while higher layers have more experts.

\paragraph{MoLA Hourglass (MoLA-$\bowtie$)}
A third model assumes that both lower and higher layers require more experts as they focus on processing basic features and abstract features. The middle layers play a role in aggregating the basic features and mapping them to a high-dimensional space for abstract reasoning, requiring fewer fine-grained features. Our MoLA Hourglass (MoLA-$\bowtie$) architecture uses this assumption to allocate experts in an ``hourglass" shape, where lower and higher layers have more experts than the middle layers. 

\paragraph{MoLA Rectangle (MoLA-$\square$)}
The last configuration is the original design of MoE, where each Transformer layer has the same number of experts. Most of the recent studies adopt this expert allocation design. We call this MoLA Rectangle (MoLA-$\square$) and use it as a baseline.

\section{Experiments}
\label{sec:experiments}

\subsection{Experiment Settings}
\label{sec:experiments_1}
We designed two experimental settings to examine the performance of our proposed MoLA approach, including instruction-tuning$\rightarrow$fine-tuning and direct fine-tuning. To make the comparisons straightforward and clear, we designed 4 allocation configurations of MoLA for the large language model as illustrated in Section~\ref{sec:MoLA_configurations}. We take LLaMA-2~\citep{touvron2023llama} which contains 32 layers as our base model. For MoLA-$\triangle$, we allocate \textbf{8} experts to each layer for the first 8 layers, \textbf{6} experts to each layer for the next 8 layers, \textbf{4} experts to each layer for 17-24 layers, and \textbf{2} experts to each layer for the last 8 layers, which is denoted as \textbf{8642}. Following the same notation, we allocate MoLA Inverted Triangle as \textbf{2468}. The allocations for Hourglass and MoLA Rectangle are \textbf{8228} and \textbf{5555}, separately. Notably, to make the comparison fair, we make the total number of experts the same for all the variants, i.e., the same number of trainable parameters. The trainable parameter number is 105,635,840, which is a 1.5\% trainable parameter number of the pre-trained base model.

In the first setting, we perform instruction tuning ~\citep{wei2021finetuned,sanh2022multitask,mishra2022cross} with PEFT methods on an instruction-tuning dataset with cross-entropy loss. We also adopt auxiliary loss for balancing the top-k selection of routing following Switch Transformers~\citep{fedus2022switch}. We then fine-tune the pre-trained PEFT models on downstream tasks. In the second setting, we directly fine-tune the PEFT models on downstream tasks without instruction tuning.

\subsection{Task and Data}
MoLA is intended to fine-tune LLMs on downstream tasks and/or fine-tune instructions. To show its effectiveness, we study both natural language processing (NLP) tasks and commonsense reasoning (CR) tasks. For NLP tasks, we evaluate three popular datasets, including Microsoft's Research Paraphrase Corpus \citep{mrpc}, Recognizing Textual Entailment (RTE) dataset \citep{glue}, and Corpus of Linguistic Acceptability (COLA) \citep{glue}. For commonsense reasoning tasks, we evaluate three recent question-answering benchmarks, including ScienceQA \citep{scienceqa}, CommonsenseQA \citep{commonsenseqa}, and OpenbookQA \citep{openbookqa}. We follow the task-specific fine-tuning framework to evaluate their effectiveness.

Microsoft's Research Paraphrase Corpus (MRPC) consists of 5,801 sentence pairs collected from newswire articles. Each pair is labeled by whether it is a paraphrase or not and the task is to classify whether sentence pairs are paraphrases. The dataset is divided into a training set with 4,076 sentence pairs and a testing set with 1,725 pairs.

The Recognizing Textual Entailment (RTE) dataset comes from a series of annual textual entailment challenges including RTE1, RET2, RTE3, and RTE5. Sentence examples are constructed based on news and Wikipedia text. This dataset is a two-class classification, entailment or not entailment, containing 2,490 training and 277 validation samples.

The Corpus of Linguistic Acceptability (COLA) consists of English acceptability judgments drawn from books and journal articles on linguistic theory. Each example is a sequence of words annotated with whether it is a grammatical English sentence. This corpus has 8,551 training samples and 1,043 validation samples for checking grammar.

ScienceQA is a commonsense question-answering dataset collected from elementary and high school science curricula containing 21,208 multimodal multiple-choice sentence questions. Because this paper focuses on textual inputs, we gathered all text-only samples and created a training and test set of 6508 and 2224 samples, respectively. ScienceQA has rich diversity from three subjects: natural science and language science, social science. To answer these questions, models need to align with correct commonsense knowledge.

CommonsenseQA is a commonsense reasoning question-answering dataset that requires different types of commonsense knowledge to predict the correct answers. The dataset was generated by Amazon Mechanical Turk workers and contains 9,740 training samples and 1,221 validation samples.

OpenbookQA is a commonsense question-answering dataset for assessing human understanding of a subject. It consists of 5,957 multiple-choice elementary-level science questions with 4,957 training, 500 validation, and 500 test samples. To answer a question, models must probe the understanding of a small ``book" of 1,326 core science facts and the application of these facts to novel situations. 

Besides all the evaluation benchmarks, we also use an instruction-tuning corpus for training. We randomly sampled 50,000 samples from the OpenOrca \citep{OpenOrca} corpus. OpenOrca is an instruction-tuning dataset consisting of augmented FLAN data aligned with the distributions demonstrated in ORCA~\citep{mukherjee2023orca} and it has 2.91M data samples across diverse tasks or instructions.

\begin{table*}[!ht]
    \caption{Comparison with different methods on directly downstream fine-tuning. MoLA-$\triangledown$ outperforms other variants or baselines and even achieves competitive or superior performance with MoLA-$\square$ (8888), with nearly 40\% fewer parameters. }
    \vskip 0.1in
	\centering	
    \small
        \begin{tabular}	{l | c  c  c|  c  c  c }
        \toprule	 	
        Models (\# of Experts) &  MRPC & COLA & RTE &  ScienceQA & CommonsenseQA & OpenbookQA \\
        \midrule
        Full-Parameter  & 87.13\% & 86.29\% & 87.73\% & 93.12\% & 77.48\% & 80.4\% \\
        \midrule
        Prompt Tuning & 49.91\% & 59.25\% & 54.17\% & 36.78\% & 37.76\% & 46.2\% \\
        LLaMA-Adapter & 71.94\% & 47.56\% & 72.93\% & 73.33\% & 73.55\% & 71.8\% \\
        LoRA & 83.13\% & \textbf{86.29\%} & 85.92\% & 91.01\%  & 75.51\% & 77.0\% \\
        MoLA-$\square$ (8888) & \textbf{84.70\%} & 85.81\% & \textbf{88.45\%} & \textbf{91.91\%} & \textbf{77.89\%} & \textbf{82.8\%}\\
        \midrule
        MoLA-$\square$ (5555) & 84.23\% & 86.28\% & 85.20\% & 92.04\% &  78.13\% & \textbf{80.0\%}\\
        MoLA-$\triangle$ (8642) & \textbf{84.64\%} & 85.43\% & 84.84\% & 91.90\% & 77.23\% & 77.6\% \\
        MoLA-$\bowtie$ (8228) & 83.48\% &  86.00\% & \textbf{86.28\%} & 91.41\%  & 76.25\% & 78.8\% \\	
        MoLA-$\triangledown$ (2468) & 83.48\% & \textbf{86.87\%} & \textbf{86.28\%} & \textbf{92.36\%}  & \textbf{78.95\%} & 79.6\% \\	
        \bottomrule	
        \end{tabular}
	\label{table1:finetuning}
\end{table*}

\begin{table*}[!ht]
    \caption{Comparison with different methods on instruction-tuning \& downstream fine-tuning. MoLA-$\triangledown$ outperforms other variants and shows promising transfer learning capability.}
    \vskip 0.1in
	\centering	
    \small
        \begin{tabular}	{l | c  c  c|  c  c  c }
        \toprule	 	
        Models (\# of Experts) &  MRPC & COLA & RTE &  ScienceQA & CommonsenseQA & OpenbookQA \\
        \midrule
        LoRA & \textbf{84.41}\% & 84.95\% & 84.48\% & 91.01\%  & 74.61\% & 76.6\% \\
        MoLA-$\square$ (8888) & 84.23\% & \textbf{85.72\%} & \textbf{87.36\%} & \textbf{92.13\%} & \textbf{77.15\%}  & \textbf{78.4\%}\\
        \midrule
        MoLA-$\square$ (5555) & 84.93\% & 84.56\% & 88.81\% & 91.73\% & 75.92\%  & 77.6\%\\
        MoLA-$\triangle$ (8642) & 84.46\% & 85.23\% & \textbf{89.17\%} & 91.41\% & 76.33\% & \textbf{78.8\%}\\
        MoLA-$\bowtie$ (8228) & 84.35\% & 84.85\% & 87.72\% & 91.41\% &  75.02\% & 77.4\% \\	
        MoLA-$\triangledown$ (2468) & \textbf{85.45\%} & \textbf{86.19\%} & \textbf{89.17\%} & \textbf{92.36\%}  & \textbf{77.15\%} & 78.4\% \\	
        \bottomrule	
        \end{tabular}
	\label{table2:instruction_tuning}
\end{table*}

\subsection{Recent Competitive Baselines}
We compare MoLA with three parameter-efficient tuning approaches, prompt tuning~\citep{lester2021power}, LLaMA-Adapter~\citep{zhang2023llama}, and LoRA\cite{lora}. We also evaluate full-parameter fine-tuning. Prompt tuning presents soft prompting concatenated to the embedding layer of the Transformer model. Soft prompts are a set of virtual tokens pre-appended to the textual prompt and passed to the LLM. During fine-tuning, the LLM is frozen and only the virtual tokens are optimized, providing a lightweight tuning approach. LLaMA-Adapter is an adaption method for LLaMA instruction tuning and has a set of learnable adaption prompts that are pre-appended to the word tokens at higher transformer layers. A zero-initialized attention mechanism with zero gating is used to inject new instructional cues into LLaMA. LoRA was briefly described in Section~\ref{sec:MoLA}. Specifically, the rank of LoRA is 64. In our evaluation, LLMs are fine-tuned on the downstream training dataset via different parameter-efficient tuning approaches. Based on the availability of test set labels, we evaluated COLA, RTE, and CommonsenseQA on their validation set and others on the test set.

\subsection{Implementation}
We use LLAMA2-7B \citep{touvron2023llama} as our base language model across all the experiments. In the first setting, we trained the PEFT model on a sampled instruction-tuning dataset for 3 epochs. In both settings, we do a grid search on the number of training epochs, including 10, 15, and 20 epochs for downstream task fine-tuning. We use AdamW~\citep{loshchilov2017decoupled} as the optimizer with a learning rate of 3e-4. The cutoff length is set to 256 following \citet{sanh2022multitask} and the batch size is 128. The rank of each LoRA expert is 8 and we adopt top-2 for the router. LoRA alpha is set to 16 and LoRA dropout is 0.05, following the default LoRA settings. We applied LoRA to four weight matrices in the self-attention module ($W_q$ , $W_k$, $W_v$ , $W_o$) and three weight matrices in the MLP module ($W_{gate}$, $W_{down}$, $W_{up}$). All experiments were conducted on the servers with eight A100-40G GPUs. 

\subsection{Results}
\label{sec:results}
\paragraph{Comparison with Baselines} Table~\ref{table1:finetuning} shows the results for the direct fine-tuning setting where each number is the accuracy (\%) for each dataset. From the table, LoRA-based approaches (LoRA and MoLA) significantly outperform prompt-tuning-based baselines (Prompt Tuning and LLaMA-Adapter). For LoRA-based methods, the original LoRA with rank 64 is used as our baseline. We first evaluate the MoLA-$\square$ with eight experts at each layer, annotated as MoLA-$\square$(8888), where the number of parameters is the same as the LoRA baseline. Then, we reduce the sum of configuration number from 32 ($8\times4$)  to 20 in total, with only 62.5\% of the parameters, and evaluate the four different configurations as described in Section~\ref{sec:experiments_1}. MoLA variants outperform the LoRA baseline on all the benchmarks. Specifically, MoLA-$\triangledown$ beats LoRA on all six datasets --- the performance improvements of MoLA-$\triangledown$ are larger on the commonsense QA tasks compared to the NLP tasks. It even outperforms the MoLA-$\square$(8888) on three benchmarks with nearly 40\% fewer parameters. The results demonstrate the effectiveness and scalability of MoLA.

Table.~\ref{table2:instruction_tuning} presents the results, in accuracy(\%), for the instruction-tuning$\rightarrow$fine-tuning setting. The language model is first tuned via each PEFT approach on our instruction-tuning set. The model is then fine-tuned on all downstream tasks. This setting evaluates the transfer learning capability of each PEFT approach. We only compare the LoRA-based methods due to their superior transfer learning capabilities (vs. prompt-tuning-based methods). Our results show that MoLA variants significantly outperform LoRA on all the datasets. We observe that instruction tuning provides more performance gains using MoLA compared with LoRA. For example, our MoLA-$\triangledown$ outperforms LoRA by 0.3 on MRPC in the direct fine-tuning setting, and this improvement increases to 1.04 in this setting. With instruction tuning, MoLA-$\triangledown$ achieves either equal or better performance compared with MoLA-$\square$(8888) on all the datasets even with much fewer parameters.

Tables ~\ref{table1:finetuning} and \ref{table2:instruction_tuning} show that MoLA-$\triangle$ and -$\bowtie$ perform worse than MoLA-$\square$ and MoLA-$\triangledown$, especially in the QA task. Of all MoLA variants, MoLA-$\triangledown$ generally achieves the best performance, outperforming all other variants on five benchmarks.  We performed the Wilcoxon signed-rank test with False Discovery Rate (FDR) correction among MoLA-$\triangledown$ and other baselines (Prompt Tuning, LLaMA-Adapter, LoRA) based on their accuracy on six benchmarks. The superior performance of MoLA-$\triangledown$ compared to other baselines was verified to be statistically significant with all p-values being less than 0.05.

\emph{These experiments show that allocating more experts in the higher layers and fewer experts in the lower layers provides better effectiveness compared with other allocation strategies.}  We therefore argue that the number of experts at the top layers is important. In other words, if we would like to prune the MoLAs, it is better to reduce the number of experts at lower layers to reduce the trainable parameters. In the next section, we explore the possible rationales behind this configuration and some other properties of MoLA.


\section{Model Analysis and Ablation Studies}
\label{sec:analysis}
\subsection{Analysis of Layer-wise Experts Redundancy}
In the previous section, we observed that allocating more LoRA experts to higher layers provides more performance gains. Thus, based on our assumptions, higher layers should be allocated more experts than the lower layers. More experts with fine-grained processing on token-level information may introduce redundancy. On the contrary, higher layers require more experts because higher layers learn more abstract and high-level information. More experts can enhance the architecture to learn fine-grained and task-specific patterns for complicated downstream problems. Here, we study the redundancy of the layer-wise LoRA experts to convince our assumptions.

\begin{figure}[htb]
\begin{center}
\centerline{\includegraphics[width=\columnwidth]{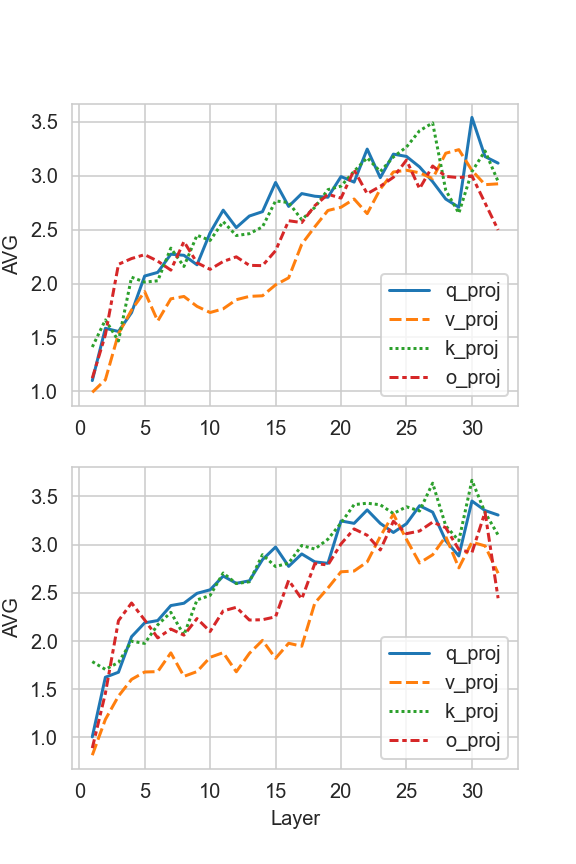}}
\caption{Average number of the Frobenius Norm between two different experts' weight matrices for each self-attention module from each layer. The top figure is for the MoLA-$\square$(8888), and the bottom figure is for MoLA-$\square$ (5555). Both models are trained via instruction tuning.}
\label{fig3:norms}
\end{center}
\vskip -0.5in
\end{figure}

To better analyze the models, we formally define the layer-wise expert redundancy as follows:
\begin{definition}
\textbf{Expert Redundancy} measures the layer-wise difference between expert modules in MoE architecture for Transformer models. 
\end{definition}
When two selected experts are similar, they may overlap and create some redundancy. To quantitatively examine the Expert Redundancy of Transformer layer $j$, we calculate the average value of the Frobenius Norm between any two different LoRA experts' weight matrices in each self-attention module from layer $j$. Figure.~\ref{fig3:norms} presents the layer-wise average values of MoLA-$\square$(8888) and MoLA-$\square$(5555), where both models are trained via the sampled instruction-tuning dataset. In the figure, the average value of the Frobenius Norm per layer increases from lower layers to higher layers, showing that experts in lower layers are more similar than those in higher layers. We also find a similar pattern from other MoLA configurations, as demonstrated in Appendix.~\ref{apx:norm}. This observation supports our assumption that experts in lower layers of the Transformer model suffer more expert redundancy. Therefore, within a certain number of experts in total, allocating more experts in higher layers is more effective in improving the model performance.

We also analyze the average router weights and selected times of experts in each layer. Analysis shows that most of the experts are selected for a similar workload and utilized sufficiently (details are described in Appendix.~\ref{apx:avg_fusion_selected}). With similar selected weights and times, lower layers have more expert redundancy due to more similar experts, again supporting our statement.

\subsection{Continuous Learning}
MoE-based architectures leverage the sparse sub-modules to process information, and thus only selected modules are optimized for different input types. This feature may also provide more stable performance for continuous learning. Here, we explore the domain continuous learning ability of our MoLAs and perform experiments on the ScienceQA dataset. We choose the 5 topics with the most training samples including biology, physics, chemistry, economics, and earth science. We continuously fine-tune MoLAs on new domains and study the performance drop on previous domains. Following \citet{chaudhry2018riemannian}, we calculate the overall performance (OP):
\begin{equation}
OP = \frac{1}{t}\sum_{i=1}^{t}R_{t,i},
\label{eq5:OP}
\end{equation}
where $t$ is the number of domain and $R_{t,i}$ denotes the model's accuracy on domain $i$ after continuously trained on domain $t$. We also propose a performance drop score to measure the domain forgetting by calculating the performance drop in the continuous learning process, as illustrated in the following equation:
\begin{equation}
PD = \frac{1}{t(t-1)/2}\sum_{k=2}^{t}\sum_{i=1}^{k-1}(R_{k,i}-R_{k-1,i}),
\label{eq6:PD}
\end{equation}
\begin{table}[!ht]
    \caption{Comparison with various MoLA in a continuous setting.}
    \vskip 0.1in
	\centering	
    \small
        \begin{tabular}	{l | c  c }
        \toprule	 	
        Models &  OP $\uparrow$ & PD $\uparrow$ \\
        \midrule
        LoRA & 78.67\% & -2.17\%\\
        MoLA-$\square$ (5555) & 88.80\% & -0.6\% \\
        MoLA-$\bowtie$ (8228) & 83.82\% & -3.92\%  \\	
        MoLA-$\triangle$ (8642) & 88.84\% & -2.10\% \\
        MoLA-$\triangledown$ (2468) & \textbf{89.82\%} & \textbf{-0.47\%} \\	
        	
        \bottomrule	
        \end{tabular}
	\label{table4:cl}
 \vskip -0.05in
\end{table}
As shown in Table~\ref{table4:cl}, MoLAs can achieve better overall performance than LoRA. Specifically, MoLA-$\triangledown$ shows the superior ability to avoid domain knowledge forgetting by having a -0.47 performance drop score, which aligns with our insights that the higher layers have less expert redundancy. The detailed results are shown in Appendix.~\ref{apx:cl}.

\section{Related Work}
\label{sec:relatedwork}
Here, we describe works most closely linked to this effort. These include work on parameter-efficient tuning and recent research combining MoE and parameter-efficient tuning.

\subsection{Parameter-Efficient Tuning} Parameter-efficient tuning of LLMs has garnered considerable attention because it is cost-effective nature for fine-tuning LLMs. \citet{softprompt} and \citet{ptuning} present the use of soft prompting concatenated to either the embedding layer or intermediate layers of the Transformer model. However, these approaches involve adding extra embedding tokens to the sequence, potentially compromising efficiency during inference, especially in the case of long input contexts. \citet{lora} introduces the LoRA parameter-efficient adaptation technique which uses low-rank decomposition matrices of dense weight matrices of Transformers. LoRA achieves decent performance for fine-tuning LLMs without additional inference costs. Similarly, \citet{ia3} uses task-specific vectors to modify attention activation, also avoiding extra inference costs. Inspired by these approaches, our approach combines the MoE technique with parameter-efficient tuning approaches and leverages the layer-wise expert allocation to further push the limit of performance.

\subsection{Parameter-Efficient MoE} Some recent efforts have studied the integration of MoE and parameter-efficient tuning methods to improve the effectiveness of instruction tuning. \citet{moelora} applies MoE with LoRA matrices for fine-tuning language models on various medical domain tasks. This method takes the task type as an additional input for training the router, which requires additional prior knowledge during inference. Our approach does not require additional prior knowledge since our MoLA experts are learned without supervision. \citet{fudanloramoe} introduced LoRAMoE, a novel adapter architecture that combines MoE and LoRA within the feed-forward layer of each Transformer block. This effort also studies how to mitigate knowledge forgetting in LLMs during traditional supervised fine-tuning. However, this paper only applies LoRAMoE on the feed-forward layer in each Transformer block. MoLA, on the other hand, applies LoRA experts across each dense weight matrix in the Transformer, further improving both the performance and scalability of parameter-efficient fine-tuning. \citet{cohereloramoe} introduces a framework that combines MoE with various parameter-efficient architectures, including LoRA and IA3 \citep{ia3}, called MoLORA and MoV. Their experiments show that their framework leverages instruction tuning more effectively than prior parameter-efficient architectures, improving the zero-shot capabilities of LLMs. However, they did not study how this framework works on decoder-only LLMs and task-specific fine-tuning. Furthermore, the previously mentioned methods do not consider the layer-wise allocation of experts. Our MoLA approach introduces a novel design that allows for a varying number of experts in each layer, therefore further improving the effectiveness of LoRA-MoE approaches.

\section{Conclusion}
\label{sec:conclusion}
We introduce MoLA, a novel parameter-efficient tuning approach that leverages layer-wise expert allocation in the MoE and combines it with the LoRA technique. We propose four layer-wise expert configurations, MoLA-$\triangle$, MoLA-$\triangledown$, MoLA-$\bowtie$, and MoLA-$\square$ based on different assumptions. Our comprehensive experiments on six popular benchmarks including NLP and commonsense question-answering tasks demonstrate that MoLA significantly outperforms other baselines. Specifically, MoLA-$\triangledown$ achieves the best performance in all the configurations, convincing our assumption that with a certain number of experts in total, higher layers need to be allocated more experts. We conduct extensive analysis to explore the layer-wise expert redundancy, observing that lower layers of the Transformer model suffer higher expert redundancy with MoLA tuning. Ablation studies also show that MoLA has promising continuous learning capability. As a plug-and-play PEFT approach, MoLA can be used on wide tasks. Furthermore, this work provides a promising research direction to enhance the MoE technique and PEFT approach. In the future, we will explore dynamic learning layer-wise expert allocation and apply this approach to more diverse tasks.
\newpage
\section*{Broader Impact}
This paper contributes to the advancement of parameter-efficient tuning within the field of machine learning. To the best of our knowledge, it does not directly raise any specific ethical concerns. However, it is important to note that our research relies on pre-trained large language models that may exhibit preferential biases~\citep{tang2023llamas}. Users of these models should be cognizant of these biases and consider their potential implications. Our approach is a plug-and-play parameter-efficient tuning method and can be used for diverse tasks. We push the performance limits of PEFT methods and provide decent performance on LLM fine-tuning while dramatically reducing training costs. We believe that this research direction will benefit energy saving and advance the decarbonization of AI. Also, efficient training efficiency and promising performance promote wider groups of people to leverage our approach on more practical problems.

\bibliography{main_submission}
\bibliographystyle{icml2024}

\newpage
\appendix
\onecolumn
\section{Continuous Learning}
\label{apx:cl}
We evaluate the performance of LoRA and our MoLAs in the domain's continuous learning setting. We fine-tuning the models on biology, physics, chemistry, economics, and earth-science domains sequentially. The training epochs are 20, and we use the same hyper-parameters as we used for direct fine-tuning. The detailed results are shown in Table.~\ref{table5:cls}, where the score of Bio-Phy denotes the result when the model is trained on the biology domain and tested on the physics domain.
\begin{table}[!ht]
    \caption{The results in the continuous learning setting. Bio-Phy denotes that the model trained on the biology domain is tested on the physics domain.}
    \vskip 0.1in
	\centering	
    \small
        \begin{tabular}	{l | c  c  c  c  c }
        \toprule	 	
         & LoRA & MoLA-$\triangle$ (5555) & MoLA-$\bowtie$ (8228) & MoLA-$\triangledown$ (8642) & MoLA-$\triangledown$ (2468) \\
        \midrule
        Bio-Bio     & 92.19\% & 94.71\% & 95.97\% & 95.97\% & 94.96\% \\
        Bio-Phy     & 61.46\% & 60.94\% & 64.06\% & 64.58\% & 66.67\% \\
        Bio-Chem    & 55.46\% & 63.87\% & 55.46\% & 57.98\% & 61.34\% \\
        Bio-Econ    & 60.71\% & 72.62\% & 66.67\% & 70.24\% & 66.67\% \\
        Bio-Earth   & 52.31\% & 63.08\% & 55.38\% & 52.31\% & 61.54\% \\
                    &       &       &       &       &       \\
        Phy-Bio     & 88.16\% & 91.44\% & 91.44\% & 92.44\% & 91.69\% \\
        Phy-Phy     & 89.58\% & 89.06\% & 92.19\% & 90.10\% & 88.54\% \\
        Phy-Chem    & 55.46\% & 36.13\% & 51.26\% & 57.14\% & 52.10\% \\
        Phy-Econ    & 72.62\% & 79.76\% & 78.57\% & 80.95\% & 77.38\% \\
        Phy-Earth   & 61.54\% & 58.46\% & 76.92\% & 66.15\% & 63.08\% \\
                    &       &       &       &       &       \\
        Chem-Bio    & 85.14\% & 90.43\% & 82.62\% & 89.67\% & 87.91\% \\
        Chem-Phy    & 81.77\% & 81.25\% & 85.42\% & 85.42\% & 91.15\% \\
        Chem-Chem   & 93.28\% & 94.96\% & 94.96\% & 95.80\% & 94.12\% \\
        Chem-Econ   & 59.52\% & 61.90\% & 58.33\% & 48.81\% & 72.62\% \\
        Chem-Earth  & 58.46\% & 60.00\% & 60.00\% & 52.31\% & 66.15\% \\
                    &       &       &       &       &       \\
        Econ-Bio    & 78.84\% & 89.67\% & 83.12\% & 88.66\% & 87.15\% \\
        Econ-Phy    & 78.13\% & 88.02\% & 83.85\% & 87.50\% & 86.98\% \\
        Econ-Chem   & 70.59\% & 92.44\% & 92.44\% & 96.64\% & 94.12\% \\
        Econ-Econ   & 73.81\% & 82.14\% & 94.05\% & 94.05\% & 86.90\% \\
        Econ-Earth  & 38.46\% & 56.92\% & 60.00\% & 47.69\% & 61.54\% \\
                    &       &       &       &       &       \\
        Earth-Bio   & 84.13\% & 88.66\% & 83.38\% & 86.15\% & 88.16\% \\
        Earth-Phy   & 77.08\% & 85.94\% & 81.77\% & 85.42\% & 91.67\% \\
        Earth-Chem  & 87.39\% & 93.28\% & 89.08\% & 94.12\% & 90.76\% \\
        Earth-Econ  & 78.57\% & 86.90\% & 83.33\% & 89.29\% & 89.29\% \\
        Earth-Earth & 66.15\% & 89.23\% & 81.54\% & 89.23\% & 89.23\% \\
        	
        \bottomrule	
        \end{tabular}
	\label{table5:cls}
\end{table}

\section{Frobenius Norms of Different Experts Allocation}
\label{apx:norm}
\begin{figure*}[htb]
\begin{center}
\centerline{\includegraphics[width=0.5\columnwidth]{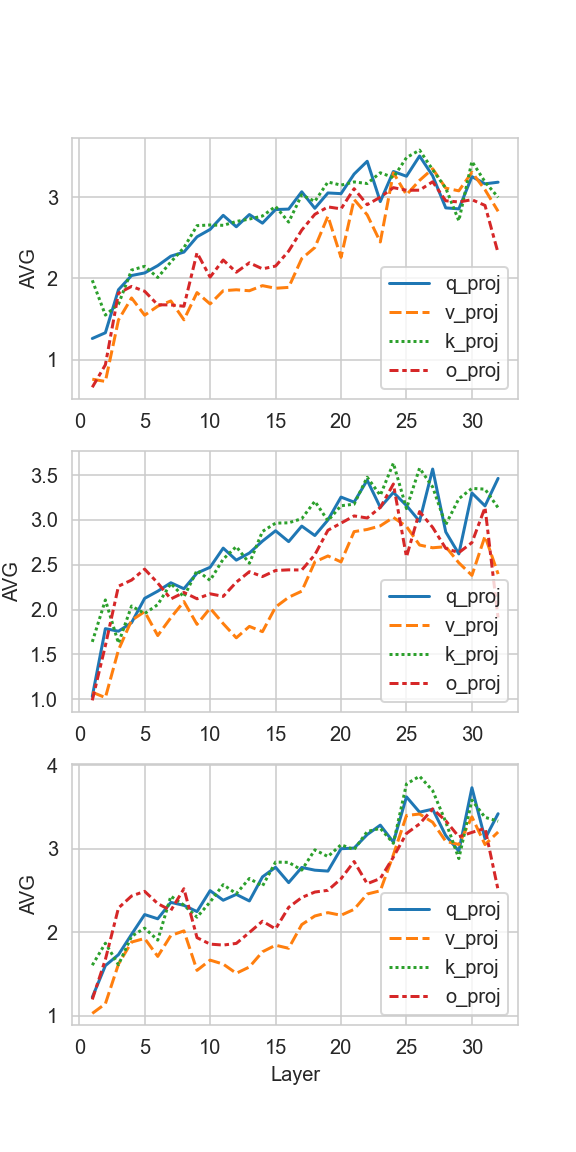}}
\caption{The average number of the Frobenius Norm between two different experts' weight matrices at the same layer for each self-attention module. The top figure is for the MoLA-$\triangledown$ with configuration as 2468; the middle figure is for the MoLA-$\triangle$ with configuration as 8642; and the bottom figure is for MoLA-$\bowtie$ with configuration 8228 after instruction tuning.}
\label{fig5:various_norms}
\end{center}
\end{figure*}

We show the Frobenius Norms for various MoLA in Figure.~\ref{fig5:various_norms}. In the figure, the top sub-figure is for the MoLA-$\triangledown$ with configuration as 2468; the middle sub-figure is for the MoLA-$\triangle$ with configuration as 8642; and the bottom sub-figure is for MoLA-$\bowtie$ with configuration 8228 after instruction tuning. All various MoLAs follow the same pattern, and the difference between the weight matrices of the experts becomes larger as the layer becomes higher.

\section{Analysis of Average Fusion Weights and Selected Times of Experts}
\label{apx:avg_fusion_selected}
\begin{figure}[!htb]
\begin{center}
\centerline{\includegraphics[width=0.5\columnwidth]{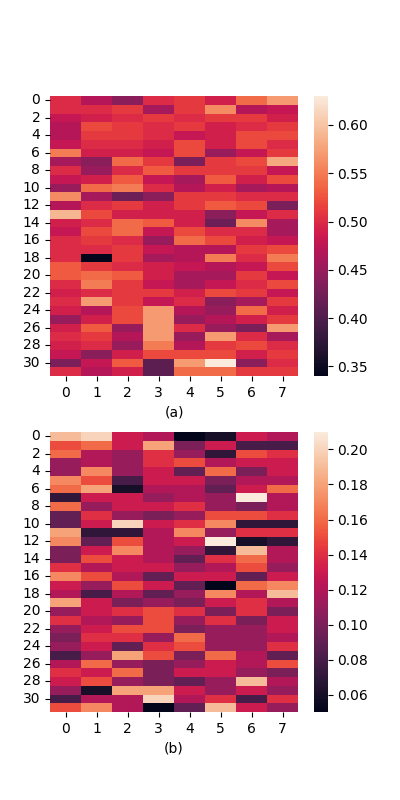}}
\caption{(a) The average fusion weights for each expert. (b) The average times for each expert when it is selected.}
\label{fig4:value_weight_selected}
\end{center}
\end{figure}

We also calculate the average fusion weights provided by the router and average times for experts when they are selected, as shown in Figure~\ref{fig4:value_weight_selected}. In Figure~\ref{fig4:value_weight_selected} (a), we find that most fusion weights are around 0.5, which means the importance of selected experts is similar most of the time. Furthermore, in Figure~\ref{fig4:value_weight_selected} (b), although there are several experts are not often selected, most of the experts are selected frequently and utilized sufficiently, which supports our insights as well. 

\end{document}